\documentclass[conference]{IEEEtran}
\IEEEoverridecommandlockouts
\usepackage{cite}
\usepackage{amsmath,amssymb,amsfonts}
\usepackage{graphicx}
\usepackage{textcomp}
\usepackage{xcolor}

\usepackage{booktabs}
\usepackage{multirow}
\usepackage{algorithm}
\usepackage{algpseudocode}
\usepackage{tabularx}
\usepackage{amsmath}
\usepackage{amsthm}
\usepackage{enumitem}
\def\BibTeX{{\rm B\kern-.05em{\sc i\kern-.025em b}\kern-.08em
    T\kern-.1667em\lower.7ex\hbox{E}\kern-.125emX}}

\newcommand{\sysname}{FEED}
\newcommand{\sysnameabs}{Abs1, w/o inner loop}
\newcommand{\sysnameabss}{Abs2, w/o augment}

\makeatletter
\newcommand{\multiline}[1]{%
  \begin{tabularx}{\dimexpr\linewidth-\ALG@thistlm}[t]{@{}X@{}}
    #1
  \end{tabularx}
}
\makeatother 

\newcommand{\argmin}{\operatornamewithlimits{argmin}}

\newtheorem{definition}{Definition}

\newtheorem{assumption}{Assumption}
\newtheorem{problem}{Problem}

\usepackage[capitalize]{cleveref}
\crefname{section}{Sec.}{Secs.}
\crefname{table}{Tab.}{Tabs.}
\crefname{assumption}{Assumps.}{Assumps.}
\crefname{problem}{problem}{problems}

\newcommand{\eqspace}{-2}

\begin{document}

\title{FEED: Fairness-Enhanced Meta-Learning for Domain Generalization\\
% {\footnotesize \textsuperscript{*}Note: Sub-titles are not captured for https://ieeexplore.ieee.org  and
}

\author{
\IEEEauthorblockN{Kai Jiang$^1$, Chen Zhao$^2$, Haoliang Wang$^1$, Feng Chen$^1$}
\IEEEauthorblockA{
$^1$\textit{Department of Computer Science, The University of Texas at Dallas, Richardson, Texas, USA} \\
$^2$\textit{Department of Computer Science, Baylor University, Waco, Texas, USA}\\
\{kai.jiang, haoliang.wang, feng.chen\}@utdallas.edu, chen\_zhao@baylor.edu}
% \and
% \IEEEauthorblockN{Chen Zhao}
% \IEEEauthorblockA{Department of Computer Science \\
% \textit{Baylor University}\\
% Waco, Texas, USA \\
% chen\_zhao@baylor.edu}
% \and
% \IEEEauthorblockN{Haoliang Wang}
% \IEEEauthorblockA{Department of Computer Science \\
% \textit{The University of Texas at Dallas}\\
% Richardson, Texas, USA \\
% haoliang.wang@utdallas.edu}
% \and
% \IEEEauthorblockN{Feng Chen}
% \IEEEauthorblockA{Department of Computer Science \\
% \textit{The University of Texas at Dallas}\\
% Richardson, Texas, USA \\
% feng.chen@utdallas.edu}
% \and
% \and
}

\maketitle

\begin{abstract}
Generalizing to out-of-distribution data with being aware of model fairness is a significant and challenging problem in meta-learning. The goal of this problem is to find a set of fairness-aware invariant parameter of classifier that is trained using data drawn from a family of related training domains with distribution shift on non-sensitive features as well as different levels of dependence between model predictions and sensitive features so that the classifier can achieve good generalization performance on unknown but distinct test domains. To tackle this challenge, existing state-of-the-art methods either address the domain generalization problem but completely ignore learning with fairness, or solely specify shifted domains with various fairness levels. This paper introduces an approach to fairness-aware meta-learning that significantly enhances domain generalization capabilities. Our framework, Fairness-Enhanced Meta-Learning for Domain Generalization (\sysname{}), disentangles latent data representations into content, style, and sensitive vectors. This disentanglement facilitates the robust generalization of machine learning models across diverse domains while adhering to fairness constraints. Unlike traditional methods that focus primarily on domain invariance or sensitivity to shifts, our model integrates a fairness-aware invariance criterion directly into the meta-learning process. This integration ensures that the learned parameters uphold fairness consistently, even when domain characteristics vary widely. We validate our approach through extensive experiments across multiple benchmarks, demonstrating not only superior performance in maintaining high accuracy and fairness but also significant improvements over existing state-of-the-art methods in domain generalization tasks.
\end{abstract}
\begin{IEEEkeywords}
Fairness-aware Meta-Learning, Domain Generalization.
\end{IEEEkeywords}

\section{Introduction}
The widespread adoption of machine learning across various sectors has underscored the critical importance of developing algorithms that can perform well across diverse domains. This challenge, often termed domain generalization, is crucial in environments that differ from the training settings, a common scenario in real-world applications such as healthcare, finance, and social justice. In these applications, not only is high accuracy essential, but fairness cannot be overlooked, especially when sensitive attributes like gender or ethnicity are involved \cite{Zhao-ICDM-2019,zhao-KDD-2024,shao2024supervised}.

Recent advancements in domain generalization techniques \cite{arjovsky2019invariant,li2018learning} have aimed at learning domain-invariant features. However, these methods often fail to address changes in distributions of sensitive attributes across domains, leading to potential fairness issues when deployed in varied real-world settings \cite{creager2021environment}.

Recent approaches in fairness-aware meta-learning have shown promise in addressing this gap by not only adapting models to new tasks with minimal data but also by potentially incorporating fairness directly into the learning process. However, existing approaches such as \cite{zhao-KDD-2021,zhao-KDD-2022,zhao-KDD-2023} focus predominantly on online learning scenarios or specific types of domain shifts, thus limiting their applicability in a broader range of domain generalization contexts.

Different from existing settings for the problem of fairness-aware domain generalization, we define the same problem but in a more general way. We illustrate our setting using the ccMNIST image dataset. In this example, data domains are specified by various digit colors, but \textit{we do not assume group fairness levels for domains are different or the same}. The goal of this problem is to learn an invariant classifier across observed training domains and achieve good generalization performance on testing domains with unknown non-sensitive variation and an unknown group fairness level.
\begin{figure}[!t]
    \centering
    \includegraphics[width=\linewidth]{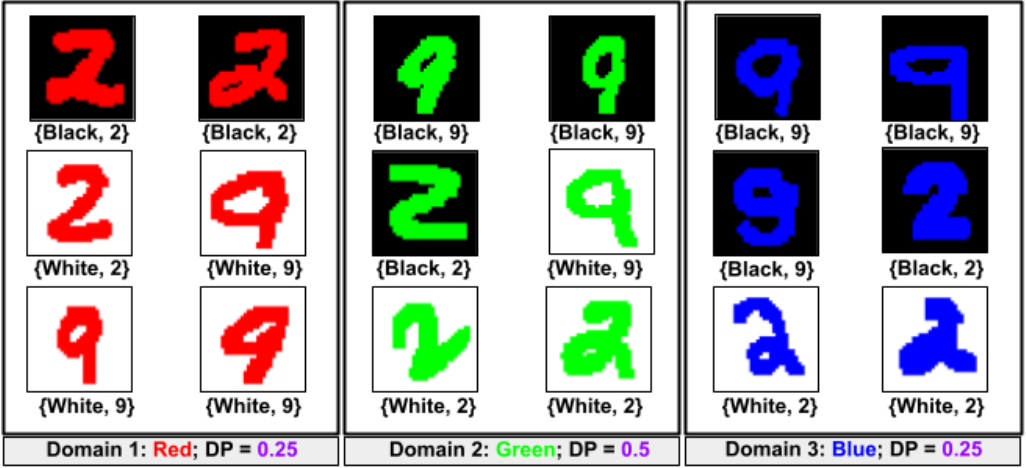}
    \vspace{-5mm}
    \caption{Illustration of fairness-aware domain generalization problems using the ccMNIST digit dataset. The domains correspond to different digit colors (red/green/blue). Each image has a black or white background color as the sensitive label. For simplicity, digits 2 and 9 are used as toy examples to demonstrate the setting. Each domain is associated with various group fairness levels estimated using the demographic parity metric.}
    \label{fig:dg-example}
    \vspace{-5mm}
\end{figure}
To address the challenges, our work introduces a novel fairness-aware meta-learning framework specifically tailored for domain generalization. This framework is designed to learn a robust set of initial parameters that are optimized for effective adaptation across a range of diverse domains, embedding fairness considerations directly at the meta-parameter level. This allows the model to rapidly adapt to new domains while adhering to stringent fairness constraints, thereby pushing the boundaries of traditional domain generalization approaches. Our main following contributions are:
\begin{itemize}[leftmargin=*]
  \item We introduce a meta-learning framework for fairness across domains, innovatively incorporating fairness at the meta-parameter level. It enables our model to maintain fairness while outperforming traditional domain generalization approaches.
  \item We formulate a fairness-aware invariance criterion for meta-learning settings. This criterion ensures that the learned initial parameters are consistent and fair across domain shifts, significantly enhancing the fairness level of machine-learning models across various unseen domains.
  \item We empirically validate our method on multiple domain generalization benchmarks, where it demonstrates the ability to maintain high accuracy and fairness. Our rigorous testing against state-of-the-art domain generalization and fairness methods highlights the critical role of initial parameters in meta-learning for achieving fairness across different domain shifts.
\end{itemize}
\section{Related Work}
\textbf{Fairness-aware domain generalization.}
Fairness considerations in domain generalization have emerged as a concern due to challenges posed by domain shifts and the unavailability of out-of-distribution (OOD) data, which are traditionally tackled by several leading techniques \cite{vapnik1999nature,arjovsky2019invariant,sagawa2019distributionally,yan2020improve,zhang2022towards,robey2021model}. These methods strive to enhance the innate generalizability of machine learning models across source domains, each characterized by distinct but potentially overlapping distributions \cite{volpi2021continual}. A prevalent approach involves aligning distributions across multiple sources to foster domain-invariant feature representations, crucial for stable pattern recognition across domains without target domain data access \cite{li2018domain,zhou2020learning}. Notably, some strategies incorporate meta-learning paradigms to acclimate the model to domain shifts during the training phase \cite{li2018learning} or use domain analytic data augmentation techniques to broaden the model's exposure to potential shifts \cite{zhou2020learning}. 

Despite these advancements, the integration of fairness into domain generalization remains scant. Most research in domain generalization \cite{zhang2022towards,robey2021model,blanchard2011generalizing}, has predominantly focused on leveraging diverse source data to uncover invariant patterns. As \cite{blanchard2011generalizing} articulates, the principal goal is to derive representations that are robust to the marginal distributions of data features, thereby eschewing reliance on target data. However, this line of inquiry largely overlooks the nuances of ensuring that fairness across varying domains. Addressing this gap could enhance the robustness and ethical alignment of models deployed in real-world settings. A recent method for disentangling sensitive attributes, as proposed by Zhao et al. (2024) \cite{zhao2024algorithmic}, focuses on learning domain-invariant parameters from training domains. These parameters are fixed and directly applied to new domains. However, the method lacks adaptability when applied to new domains with only a few examples. These methods optimize parameters for multiple domains but are limited in rapidly adapting to unseen tasks. In contrast, our method, based on meta-learning, learns initial parameters that quickly adapt to new domains while ensuring fairness.

\textbf{Fairness-aware meta-learning.}
In the context of fairness-aware meta-learning, research efforts primarily focus on developing adaptable frameworks that can effectively handle shifts in domain characteristics while maintaining fairness standards. Strategies such as equality-aware monitoring \cite{alonso2021ordering} have been developed. These approaches continuously observe the outputs of a model to detect any deviations from fairness norms and adjust accordingly by modifying the model’s parameters or its structure. However, these methods traditionally operate under the assumption that fairness metrics remain consistent across different domains, an assumption often contradicted by the complexities encountered in practical scenarios. Zeng et al.\cite{zeng2024fairness} introduced a Nash Bargaining solution to enhance fairness in meta-learning models. However, their approach sometimes struggled with the robustness of fairness across drastic domain shifts due to an overemphasis on bargaining outcomes in homogeneous domains. In contrast, our framework enhances domain generalization by disentangling latent representations into content, style, and sensitive factors, thereby maintaining fairness even when domain characteristics vary significantly. Furthermore, alternative approaches in the literature \cite{oh2022learning,creager2021environment} attempt to evaluate a model’s fairness by recognizing changes in fairness benchmarks as indicative of domain shifts, yet they tend to overlook variations in the distribution of non-sensitive attributes, which can lead to inadequate generalization capabilities.

To address these challenges, our meta-learning framework innovatively partitions data attributes into sensitive and non-sensitive categories. Such a distinction is pivotal for the meta-learning algorithm, which is designed not merely to react to explicit domain labels but also to respond to more nuanced shifts in the distributions of data features.  This approach enables our meta-learning algorithm to refine its strategy for learning initial parameters, ensuring domain generalization and fairness. By effectively distinguishing between these attribute categories, the algorithm can prioritize the learning of initial parameters that maintain high performance and fairness standards across a spectrum of environments.
\section{Preliminaries}
\textbf{Notations.}
Consider the data space $\mathcal{P=X\times Z\times Y}$, where $\mathcal{X} \subseteq \mathbb{R}^d$ denotes a feature space, $\mathcal{Z} \subseteq \{-1, 1\}$ denotes binary sensitive attributes\footnote{In this study, we focus on a single binary sensitive attribute for clarity. Extensions to multiple sensitive attributes of various categories can be seamlessly integrated.}, and $\mathcal{Y} \subseteq \{0, 1\}$ denotes the binary output space for binary classification. Define the parameterized latent spaces: $\mathcal{C}$ for content, $\mathcal{S}$ for style, and $\mathcal{A}$ for sensitivity factors.

The function $d(\cdot,\cdot)$ is a distance measure across the space $\mathcal{Y}\times\mathcal{Y}$. Variables and parameters in our framework are symbolically denoted as follows: vectors in boldface lowercase letters, and scalars in italic lowercase letters.

\textbf{Problem setting.}
Given a dataset $\mathcal{D}$, we consider a set of data domains $\mathcal{E}=\{e_i\}_{i=1}^n$ where each domain corresponds to a distinct data subset $\mathcal{D}^{e_i}={(\mathbf{x}^{e_i}_j, z^{e_i}_j, y^{e_i}_j)}_{j=1}^{|\mathcal{D}^{e_i}|}$ over $\mathcal{P}$, and $\mathcal{D}=\bigcup_{i=1,\cdots,n} \mathcal{D}^{e_i}$.
Data domains are partitioned into multiple training domains $\mathcal{E}_{train}\subsetneq\mathcal{E}$ and testing domains $\mathcal{E}_{test} = \mathcal{E} \setminus \mathcal{E}_{train}$. The corresponding datasets are $\mathcal{D}_{train}=\bigcup_{i=1,\cdots,n_{tr}} \mathcal{D}^{e_i}$ where ${e_i}\in\mathcal{E}_{train}$ and $\mathcal{D}_{test}=\bigcup_{i=1,\cdots,n_{te}} \mathcal{D}^{e_i}$ where ${e_i}\in\mathcal{E}_{test}$.
Given samples from finite training domains $\mathcal{E}_{train}$, the goal of fairness-aware domain generalization problems is to learn initial parameters $\boldsymbol{\theta}\in\Theta$ of classifier $f$ that is generalizable across all possible domains.

\textbf{Meta-learning.} Define task $\mathcal{T}\sim p(\mathcal{T})$ where $ p(\mathcal{T})=\{ (\mathcal{B}^{sup}, \mathcal{B}^{qry}) \mid \mathcal{B}^{sup} \cup \mathcal{B}^{qry} \subseteq \mathcal{D}_{train}, \mathcal{B}^{sup} \cap \mathcal{B}^{qry} = \emptyset\}$. The goal of meta-learning is to learn initial parameters on the training dataset, and it can be quickly adapted to the testing dataset (understream task).

Model-agnostic meta-learning (MAML)\cite{finn2017model}, as a state-of-the-art approach in the meta-learning landscape, exemplifies a robust framework designed for such tasks. In MAML, a unique set of model parameters $\boldsymbol{\theta}'$ is trained for each task $\mathcal{T}$ on its support set $\mathcal{B}^{sup}$. These parameters are specifically adapted from a shared set of meta-parameters $\boldsymbol{\theta}$, which are iteratively updated based on the aggregate loss observed across all query sets $\mathcal{B}^{qry}$.

{\vspace{\eqspace mm}\small
\begin{equation}
    \boldsymbol{\theta}' = \boldsymbol{\theta} - \alpha \nabla_{\boldsymbol{\theta}} \mathcal{L}_{\mathcal{B}^{sup}}(\boldsymbol{\theta}),
\end{equation}
}
{\vspace{\eqspace mm}\small
\begin{equation}
    \boldsymbol{\theta} \leftarrow \boldsymbol{\theta} - \beta \nabla_{\boldsymbol{\theta}} \sum_{\mathcal{T} \sim p(\mathcal{T})} \mathcal{L}_{\mathcal{B}^{qry}}(\boldsymbol{\theta}')
\end{equation}
}
where $\alpha$ is the task-specific learning rate and $\beta$ denotes the meta-learning rate. $\mathcal{L}_{\mathcal{B}^{sup}}(\boldsymbol{\theta})$ represents the loss calculated on the support set using the initial meta-parameters $\boldsymbol{\theta}$, and $\mathcal{L}_{\mathcal{B}^{qry}}(\boldsymbol{\theta}')$ denotes the loss calculated on the query set using the parameters $\boldsymbol{\theta}'$.
\subsection{Assumptions}
\label{sec:prob_assumptions}
\begin{figure}[!t]
    \centering
    \includegraphics[width=0.4\linewidth]{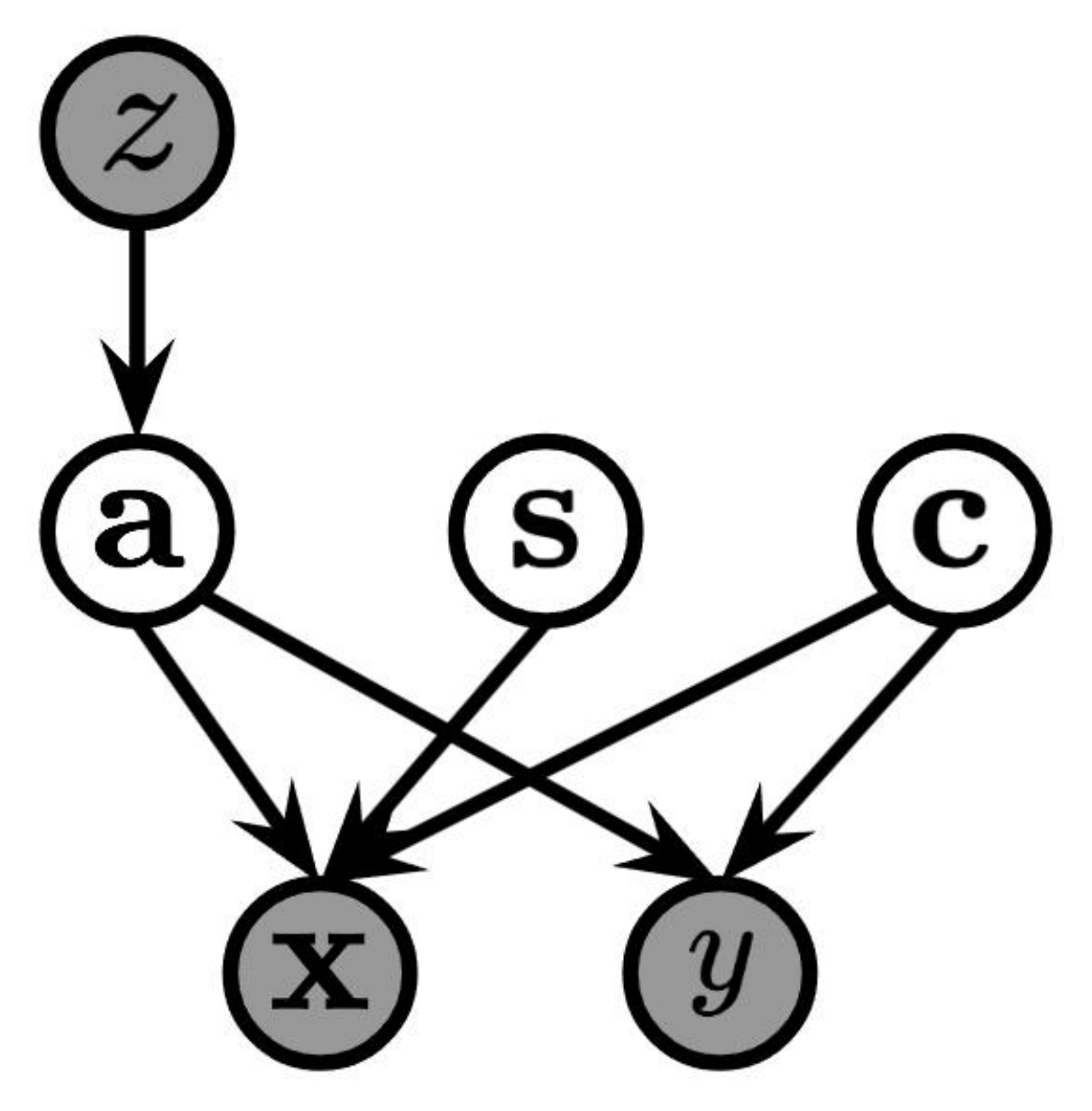}
    \caption{Causal interpretation of fairness-aware domain generalization tasks. We assume that the raw features ($\mathbf{x}$) and class label ($y$) of each example are generated by the latent content factor ($\mathbf{c}$), style factor ($\mathbf{s}$), and sensitive factor ($\mathbf{a}$). The sensitive factor  ($\mathbf{a}$) is dependent on the sensitive attribute ($z$) of this example and may or may not be dependent on the domain. The style factor $\mathbf{s}$ depends on the domain, but the content factor $\mathbf{c}$ is independent of the domain $e$. Each domain label is unobserved.}
    \label{fig:causal-graph}
    \vspace{-7mm}
\end{figure}

\begin{assumption}[Latent Spaces]
\label{assump:partially-shared-latent-spaces}
Given a batch $\mathcal{B}=\{(\mathbf{x}^{e_i}_j, z^{e_i}_j, y^{e_i}_j)\}_{j=1}^{|\mathcal{B}|}$ sampled from a specific domain $e_i \in \mathcal{E}$, as illustrated in \cref{fig:causal-graph}, we postulate that each data point $\mathbf{x}^{e_i}_j$ within the task originates from:
\begin{itemize}[]
    \item a latent content factor $\mathbf{c} \in \mathcal{C}$, where $\mathcal{C}$ denotes a content space that is invariant across all domains $\mathcal{E}$;
    \item a latent style factor $\mathbf{s} \in \mathcal{S}$ that is unique to the specific domain $e_i$;
    \item a latent sensitive factor $\mathbf{a} \in \mathcal{A}$.
\end{itemize}
where $\mathcal{C} \cap \mathcal{S} \cap \mathcal{A} = \emptyset$. Each domain $e_i$ is uniquely characterized by its style factors, denoted as $e_i := \bf s$.  
\end{assumption}
\cref{assump:partially-shared-latent-spaces} echoes the assumptions made in prior works such as \cite{zhang2022towards,robey2021model,huang2018multimodal,liu2017unsupervised}. Specifically, UNIT \cite{liu2017unsupervised} hypothesizes a fully shared latent space across all factors, whereas MUNIT \cite{huang2018multimodal} suggests a hybrid latent space model where some components are shared across domains and others are domain-specific. In our framework, considering group fairness, we extend these concepts to include three distinct latent spaces: a content space $\mathcal{C}$, a style space $\mathcal{S}$, and a sensitive space $\mathcal{A}$. Additionally, we posit that domain labels are typically unattainable in both training and testing phases due to practical limitations or excessive costs, as supported by \cite{hashimoto2018fairness}.

It is essential for fairness that the labels remain independent of variations across domains. This requirement translates to a scenario where instance conditional distributions $\{\mathbb{P}(Y^{e_i}|X^{e_i},Z^{e_i})\}_{{e_i}\in\mathcal{E}}$ differ by domain, reflective of inherent domain-specific characteristics. Within the context of this research, we posit that differences across domains, termed as domain shifts, are governed exclusively by a transformation model $T:\mathcal{X\times Z\times E}\rightarrow\mathcal{X\times Z}$. Specifically, if two samples $(\mathbf{x}^{e_i}, z^{e_i})$ and $(\mathbf{x}^{e_j}, z^{e_j})$ from different domains $e_i, e_j \in \mathcal{E}$, where $i \neq j$, exhibit identical content factors, then the sample from domain $e_j$ can be reconstructed from the sample of domain $e_i$ using the transformation $T$. This process involves $T$ extracting the invariant content from $(\mathbf{x}^{e_i}, z^{e_i})$ and subsequently applying domain-specific style and sensitivity information encoded in $e_j$ to regenerate $(\mathbf{x}^{e_j}, z^{e_j})$.
\begin{assumption}[Fairness-aware Domain Invariance]
\label{assump:FDS}
We hypothesize that the variations observed between domains are primarily driven by changes in the marginal distributions $\mathbb{P}(X^{e})$ and $\mathbb{P}(Z^{e})$ for each domain $e \in \mathcal{E}$. Consequently, we posit that the conditional distribution $\mathbb{P}(Y^{e}|X^{e},Z^{e})$ remains consistent across different domains. With a domain transformation function $T$, we assert that for any feature vector $\mathbf{x} \in \mathcal{X}$, sensitive attribute $z \in \mathcal{Z}$, and class label $y \in \mathcal{Y}$:

{\vspace{\eqspace mm}\small
\begin{equation*}
\begin{aligned}
    &\mathbb{P}(Y^{e_i}=y|X^{e_i}=\mathbf{x}^{e_i},Z^{e_i}=z^{e_i}) = \mathbb{P}(Y^{e_j}=y|(X^{e_j},Z^{e_j})\\
    & =T(\mathbf{x}^{e_i},z^{e_i},e_j)) \quad \forall e_i,e_j\in\mathcal{E}, i\neq j
\end{aligned}
\end{equation*}
}
\end{assumption}
In relation to existing literature, Robey et al. \cite{robey2021model} describe a version of $T$ that incorporates content and style factors, but overlooks the sensitive factors which are crucial for ensuring fairness in domain generalization. The domain shift driven by $T$ effectively represents how the distinct distributions $\mathbb{P}(X^{e_i})$ and $\mathbb{P}(Z^{e_i})$ map to the corresponding distributions $\mathbb{P}(X^{e_j})$ and $\mathbb{P}(Z^{e_j})$ in domains. Moreover, it is fundamental in our framework that class labels $y\sim Y$ should remain invariant to changes in fairness-sensitive attributes across domains. In this context, inter-domain variation is exclusively defined by the transformations dictated by $T$.

Our approach delves into the domain generalization problem, where inter-domain variability is specifically attributed to domain shifts driven by $T$, representing environmental discrepancies across a collection of marginal distributions $\{\mathbb{P}(X^{e_i}), \mathbb{P}(Z^{e_i})\}_{e_i\in\mathcal{E}}$. Following the assumptions set in \cref{assump:FDS}, the generation of data within each domain $e_i \in \mathcal{E}$ is conceptualized through a transformation model $T$.

To address the challenges of domain-specific variation, our methodology introduces a rigorous definition of invariance, predicated on maintaining fairness across domains as defined by the transformation model $T$.
\begin{definition}[Fairness-aware $T$-Invariance]
\label{def:T-invariance}
Let $T$ denote the domain transformation model under which a set of classifier parameters $\boldsymbol{\theta}\in\Theta$ is evaluated. A classifier is deemed fairness-aware and domain invariant if:

{\vspace{\eqspace mm}\small
\begin{align*}
    & f(\mathbf{x}^{e_i}, \boldsymbol{\theta}) = f(\mathbf{x}^{e_j},\boldsymbol{\theta}), \:\text{and} \\
    & \mathbb{E}_{\mathbb{P}(X^{e_i},Z^{e_i}),\mathbb{P}(X^{e_j},Z^{e_j})} \big[g(X^{e_i},Z^{e_i})+g(X^{e_j},Z^{e_j})\big]=0 
\end{align*}
}
is satisfied almost surely, where $(\mathbf{x}^{e_j},z^{e_j})=T(\mathbf{x}^{e_i},z^{e_i},e_j)$, $\mathbf{x}^{e_i}\sim\mathbb{P}(X^{e_i})$, $\mathbf{x}^{e_j}\sim\mathbb{P}(X^{e_j})$, and $e_i,e_j\in\mathcal{E}$.
\end{definition}
\cref{def:T-invariance} establishes the groundwork for ensuring that predictions by $f$ remain consistent across transformations induced by $T$, affirming the model's adherence to group fairness principles. The intent is that $f$ should uniformly return equivalent predictions for any data instances transformed under $T$, thereby ensuring the fairness in domain generalization.
\section{Methodology}
\subsection{Disentanglement for Fairness-aware Domain Generalization}
In our approach to enhance fairness in domain generalization, we leverage a disentanglement strategy. This strategy decomposes the samples into three distinct components: content, style, and sensitive vectors. These content vectors capture domain-invariant features essential for prediction performance, while the style vector encapsulates domain-specific variations that are irrelevant to the labels. The sensitive vector captures the sensitive attributes that could potentially lead to bias. Each sample is decomposed into these three latent vectors, enabling the generation of new samples in a synthetic domain by replacing the style and sensitive vectors with sampled ones, independent of the original domain characteristics. It allows the exploration of a more extensive and varied synthetic domain space, potentially uncovering and mitigating unfair biases that were not explicit in the original data distribution.

% We use a transformation model to transfer a sample to a new sample in a synthetic domain by utilizing 
% encoders $E^m, E^c$ and decoders ${G^i, G^o}$, which are parameterized by ${\boldsymbol{\theta}_m, \boldsymbol{\theta}_c} \in \Theta$ and ${\boldsymbol{\phi}_i, \boldsymbol{\phi}_o} \in \Phi$ respectively. Specifically, when transferring a datapoint to a new datapoint in a synthetic domain, the datapoint is first encoded to a semantic factor $\mathbf{m}\in\mathcal{M}$ through the semantic encoder $E^m:\mathcal{X}\times\Theta\rightarrow\mathcal{M}$. The semantic factor $\mathbf{m}$ is further encoded to a content factor $\mathbf{c}\in\mathcal{C}$ through $E^c:\mathcal{M}\times\Theta\rightarrow\mathcal{C}$. After sampling a sensitive factor $\mathbf{a}\in\mathcal{A}$ and a style factor $\mathbf{s}\in\mathcal{S}$, two decoders $G^{i}:\mathcal{C}\times\mathcal{A}\times\Phi\rightarrow\mathcal{M}$ and $G^{o}:\mathcal{M}\times\mathcal{S}\times\Phi\rightarrow\mathcal{X}$ are used for generating a new sample in a synthetic domain.
%Details of learning the transformation model are introduced in \cref{sec:train-trans}.
Our proposed framework involves disentangling an input sample from training domains into three factors in distinct latent spaces, using 
a series of encoders $E=\{E^m, E^s, E^c, E^a\}$ and decoders $G=\{G^i, G^o\}$. These are parameterized respectively by ${\boldsymbol{\theta}_m, \boldsymbol{\theta}_s, \boldsymbol{\theta}_c, \boldsymbol{\theta}_a} \in \Theta$ and ${\boldsymbol{\phi}_i, \boldsymbol{\phi}_o} \in \Phi$. The framework operates through two hierarchical levels: an outer level and an inner level, each with its own auto-encoder.

In the outer level, an input datapoint undergoes encoding into a semantic factor $\mathbf{m} \in \mathcal{M}$ and a style factor $\mathbf{s} \in \mathcal{S}$, achieved via the encoders $E^m: \mathcal{X} \times \Theta \rightarrow \mathcal{M}$ and $E^s: \mathcal{X} \times \Theta \rightarrow \mathcal{S}$. Progressing to the inner level, the semantic factor $\mathbf{m}$ is further decomposed into a content factor $\mathbf{c} \in \mathcal{C}$ and a sensitive factor $\mathbf{a} \in \mathcal{A}$ through the encoders $E^c: \mathcal{M} \times \Theta \rightarrow \mathcal{C}$ and $E^a: \mathcal{M} \times \Theta \rightarrow \mathcal{A}$. The corresponding decoders in these levels are $G^{i}: \mathcal{C} \times \mathcal{A} \times \Phi \rightarrow \mathcal{M}$ for the inner level and $G^{o}: \mathcal{M} \times \mathcal{S} \times \Phi \rightarrow \mathcal{X}$ for the outer level, facilitating the reconstruction of the original data.
Inspired by image-to-image translation in computer vision \cite{huang2018multimodal,liu2017unsupervised}, 
Our total loss function of learning such encoders and decoders comprises three components: a bidirectional reconstruction loss, a sensitive label prediction loss, and an adversarial loss.

% \textbf{Bidirectional Reconstruction Loss} encourages learning reconstruction in two directions: (1) data$\rightarrow$latent$\rightarrow$data for data reconstruction, and (2) latent$\rightarrow$data$\rightarrow$latent for the reconstruction of encoded factors. For simplicity, we omit parameters for encoders and decoders in the following equations.

\textbf{Reconstruction loss}
Considering a datapoint $\mathbf{x}$ sampled from $p(\mathbf{x})$, encoders and decoders in outer loop are able to reconstruct it by minimizing the reconstruction loss:

{\vspace{\eqspace mm}\small
\begin{align*}
\label{eq:x_recon}
    \mathcal{L}^x_{recon} = \mathbb{E}_{\mathbf{x}\sim p(\mathbf{x})} \left[\left\lVert G^o\left(\hat{\mathbf{m}}, E^s(\mathbf{x})\right)-\mathbf{x} \right\rVert_1\right]
\end{align*}
}
where $\hat{\mathbf{m}} = G^i(\mathbf{c,a}) = G^i\left(E^c(E^m(\mathbf{x})), E^a(E^m(\mathbf{x}))\right)$. For the inner level, the semantic factor $\mathbf{m}=E^m(\mathbf{x})$ encoded from the outer level is required to be reconstructed:

{\vspace{\eqspace mm}\small
\begin{align*}
    \mathcal{L}^{m_d}_{recon} = \mathbb{E}_{\mathbf{m}\sim p(\mathbf{m})} \left[\left\lVert G^i\left(E^c(\mathbf{m}), E^a(\mathbf{m})\right)-\mathbf{m} \right\rVert_1\right]
\end{align*}
}
with $p(\mathbf{m})$ determined by the mapping $\mathbf{m} = E^m(\mathbf{x})$ and $\mathbf{x} \sim p(\mathbf{x})$.

The latent factors $\mathbf{c, s, a}$, extracted from the datapoint $\mathbf{x}$ are encouraged to be reconstructed through some latent factors randomly sampled from the prior distributions.

{\vspace{\eqspace mm}\small
\begin{align*}
    \mathcal{L}^c_{recon} &= \mathbb{E}_{\mathbf{c}\sim p(\mathbf{c}),\mathbf{a}\sim \mathcal{N}(0,\mathbf{I}_a)} \left[\left\lVert E^c\left(G^i(\mathbf{c,a}) \right)-\mathbf{c} \right\rVert_1\right] \\
    \mathcal{L}^a_{recon} &= \mathbb{E}_{\mathbf{c}\sim p(\mathbf{c}),\mathbf{a}\sim \mathcal{N}(0,\mathbf{I}_a)} \left[\left\lVert E^a\left(G^i(\mathbf{c,a}) \right)-\mathbf{a} \right\rVert_1\right]
\end{align*}
}
where $p(\mathbf{c})$ is given by $\mathbf{c}=E^c(E^m(\mathbf{x}))$, and $\mathbf{a}=E^a(E^m(\mathbf{x}))$. Considering the dual-role of $\mathbf{m}$, as both a latent factor from the inner level and an input to the outer level, $\mathbf{s}$ can be reconstructed by two reconstruction losses:

{\vspace{\eqspace mm}\small
\begin{align*}
    \mathcal{L}^{s_{in}}_{recon} & = \mathbb{E}_{\mathbf{m}\sim p(\mathbf{m}), \mathbf{s}\sim \mathcal{N}(0,\mathbf{I}_s)} \left[\left\lVert E^s(G^o(\mathbf{m,s}))-\mathbf{s}\right\rVert_1 \right] \\
    \mathcal{L}^{s_{out}}_{recon} & = \mathbb{E}_{\mathbf{c}\sim p(\mathbf{c}),\mathbf{s}\sim \mathcal{N}(0,\mathbf{I}_s), \mathbf{a}\sim \mathcal{N}(\mathbf{0},\mathbf{I}_a)} [\lVert E^s\left(G^o(G^i(\mathbf{c,a}), \mathbf{s}) \right)\\
    &-\mathbf{s} \rVert_1]
\end{align*}
}
and for reconstructing $\mathbf{m}$ as a latent factor:

{\vspace{\eqspace mm}\small
\begin{align*}
    \mathcal{L}^{m_f}_{recon} &= \mathbb{E}_{\mathbf{m}\sim p(\mathbf{m}),\mathbf{s}\sim \mathcal{N}(0,\mathbf{I}_s)} \left[\left\lVert E^m\left(G^o(\mathbf{m,s}) \right)-\mathbf{m} \right\rVert_1\right]
\end{align*}
}

The reconstruction loss is defined as follows:

{\vspace{\eqspace mm}\small
\begin{align*}
    \mathcal{L}_{recon} &= \mathcal{L}^x_{recon} + \mathcal{L}^{m_d}_{recon} + \mathcal{L}^c_{recon} + \mathcal{L}^a_{recon} \\
    &+ \mathcal{L}^{s_{in}}_{recon} + \mathcal{L}^{s_{out}}_{recon} + \mathcal{L}^{m_f}_{recon}
\end{align*}
}

\textbf{Sensitive prediction loss} The sensitive attributes encoded from the datapoint $\mathbf{x}$ underpin the training of a classifier $h: \mathcal{A} \times \Theta \rightarrow \mathcal{Z}$. This classifier is then employed to predict the sensitive label associated with the attribute vector $\mathbf{a}$. Specifically, the prediction is formulated as:

{\vspace{\eqspace mm}\small
\begin{align*}
    & \hat{\mathbf{z}} = h(\mathbf{a}, \boldsymbol{\theta}_z) = h(E^a(E^m(\mathbf{x})), \boldsymbol{\theta}_z),\quad \mathcal{L}^z_{cls} = \text{CrossEntropy}(\mathbf{z}, \hat{\mathbf{z}})
\end{align*}
}

\textbf{Adversarial loss} Inspired by the effectiveness of Generative Adversarial Networks (GANs) \cite{goodfellow2020generative},
%in augmenting data for evaluating disentanglement in latent spaces, we employ GANs to align the distribution of reconstructed data with the original data distribution. As outlined by \cite{huang2018multimodal}, data synthesized through our encoders and decoders should be indistinguishable from actual data within the same domain.
define discriminators $D=\{D^i,D^o\}$, where $D^o:\mathcal{X}\times\Psi\rightarrow\mathbb{R}$ is the discriminator for the outer level, parameterized by $\boldsymbol{\psi}_o\in\Psi$, and $D^i:\mathcal{M}\times\Psi\rightarrow\mathbb{R}$ is the discriminator for the inner level, parameterized by $\boldsymbol{\psi}_i\in\Psi$. The discriminators are tasked with differentiating between real and constructed data with random factors.

{\vspace{\eqspace mm}\small
\begin{equation*}
\begin{aligned}
    & \mathcal{L}^x_{GAN} = \mathbb{E}_{\mathbf{c}\sim p(\mathbf{c}),\mathbf{s}\sim \mathcal{N}(0,\mathbf{I}_s), \mathbf{a}\sim \mathcal{N}(\mathbf{0},\mathbf{I}_a)} \big[\log\big(1-D^o(G^o(\hat{\mathbf{m}},\mathbf{s})) \big) \big] \\
    &+ \mathbb{E}_{\mathbf{x}\sim p(\mathbf{x})} \big[\log D^o(\mathbf{x}) \big] +\mathbb{E}_{\mathbf{c}\sim p(\mathbf{c}),\mathbf{s}\sim p(\mathbf{s}), \mathbf{a}\sim \mathcal{N}(\mathbf{0},\mathbf{I}_a)} \big[\log\big(1\\
    &-D^o(G^o(\hat{\mathbf{m}},\mathbf{s})) \big) \big] + \mathbb{E}_{\mathbf{x}\sim p(\mathbf{x})} \big[\log D^o(\mathbf{x}) \big]\\ &+\mathbb{E}_{\mathbf{c}\sim p(\mathbf{c}),\mathbf{s}\sim \mathcal{N}(0,\mathbf{I}_s), \mathbf{a}\sim p(\mathbf{a}))} \big[\log\big(1 \\
    &-D^o(G^o(\hat{\mathbf{m}},\mathbf{s})) \big) \big] + \mathbb{E}_{\mathbf{x}\sim p(\mathbf{x})} \big[\log D^o(\mathbf{x}) \big]
\end{aligned}
\end{equation*}
}
where $\hat{\mathbf{m}}$ is as defined in $\mathcal{L}^x_{recon}$.

{\vspace{\eqspace mm}\small
\begin{equation*}
\begin{aligned}
    \mathcal{L}^m_{GAN} = & \mathbb{E}_{\mathbf{c}\sim p(\mathbf{c}), \mathbf{a}\sim \mathcal{N}(\mathbf{0},\mathbf{I}_a)} \big[\log\big(1-D^i(G^i(\mathbf{c,a})) \big) \big] \\
    &+ \mathbb{E}_{\mathbf{m}\sim p(\mathbf{m})} \big[\log D^i(\mathbf{m}) \big] 
\end{aligned}
\end{equation*}
}

The adversarial loss is defined as:
% Similarly, the inner level's semantic factors generated with random sensitive attributes should also be indistinguishable from their encoded counterparts:

{\vspace{\eqspace mm}\small
\begin{equation*}
\begin{aligned}
    \mathcal{L}_{GAN} = \mathcal{L}^x_{GAN} + \mathcal{L}^m_{GAN}
\end{aligned}
\end{equation*}
}

\textbf{Total loss} We jointly train the encoders, decoders, and discriminators to optimize the final objective:
%The training of the encoders, decoders, and discriminators is orchestrated through a joint optimization of the overall objective, which integrates both adversarial and bidirectional reconstruction losses. This is expressed mathematically as:

{\vspace{\eqspace mm}\small
\begin{align*}
% \label{eq:total-loss}
    \min_{E,G}\: \max_{D} \: \mathcal{L}_{total}(E,G,D) = \mathcal{L}_{recon} + \beta_z\mathcal{L}^z_{cls} + \beta_g\mathcal{L}_{GAN}
\end{align*}
}
The $\beta_z, \beta_g > 0$ modulate the relative significance of each loss term within this formula.

This disentanglement allows the exploration of a more extensive and varied synthetic domain space, potentially uncovering and mitigating unfair biases that were not explicit in the original data distribution. Through this disentanglement, we aim to enhance the parameters' ability to generalize across domains by learning from a richer and more diverse synthetic domain data. It ensures that the learned parameters exhibit robustness to domain shifts and maintain fairness by not carrying over or amplifying biases inherent in the original data.
\subsection{Fairness-aware Meta-Learning}
\begin{problem}[Meta-Learning for Fairness-aware Domain Generalization]
\label{prob:our-problem}
Given the definitions and assumptions under \cref{def:T-invariance,assump:FDS} and a loss function $\ell:\mathcal{Y\times Y}\rightarrow\mathbb{R}$, we define the meta-learning problem as follows:

{\vspace{\eqspace mm}\small
\begin{align}
\label{eq:meta-objective}
    &\boldsymbol{\theta}^{*}=\argmin_{\boldsymbol{\theta}}  \sum_{e_i \in \mathcal{E}_{train}} \mathbb{E}_{\mathbb{P}(X^{e_i}, Z^{e_i}, Y^{e_i})} \ell(f(X^{e_i}, \boldsymbol{\theta}^{e_i}), Y^{e_i}) \\
    &\text{subject to}\quad  f(X^{e_i}, \boldsymbol{\theta}) = f(T(X^{e_i}, Z^{e_i}, e_j), \boldsymbol{\theta}), \nonumber \\
    & \mathbb{E}_{\mathbb{P}(X^{e_i}, Z^{e_i}), \mathbb{P}(X^{e_j}, Z^{e_j})} \big[g(X^{e_i}, Z^{e_i}) + g(X^{e_j}, Z^{e_j})\big] = 0 \nonumber
\end{align}
}
where the inner loop problem is defined as:

{\vspace{\eqspace mm}\small
\begin{align}
\label{eq:domain-specific-optimization}
    &\boldsymbol{\theta}^{e_i} = \argmin_{\boldsymbol{\theta}'}  \mathbb{E}_{\mathbb{P}(X^{e_i}, Z^{e_i}, Y^{e_i})} \ell(f(X^{e_i}, \boldsymbol{\theta}'), Y^{e_i}) \\
    &\text{subject to}\quad  f(X^{e_i}, \boldsymbol{\theta}') = f(T(X^{e_i}, Z^{e_i}, e_j), \boldsymbol{\theta}'), \nonumber \\
    & \mathbb{E}_{\mathbb{P}(X^{e_i}, Z^{e_i}), \mathbb{P}(X^{e_j}, Z^{e_j})} \big[g(X^{e_i}, Z^{e_i}) + g(X^{e_j}, Z^{e_j})\big] = 0 \nonumber
\end{align}
}
where $\mathbf{x}^{e_i}\sim\mathbb{P}(X^{e_i})$, $\mathbf{x}^{e_j}\sim\mathbb{P}(X^{e_j})$, $\mathbf{z}^{e_i}\sim\mathbb{P}(Z^{e_i})$, $\forall e_i,e_j\in\mathcal{E}_{train}$, $i\neq j$. $\boldsymbol{\theta}$ is the initialization of $\boldsymbol{\theta}'$ in $\boldsymbol{\theta}^{e_i}$.

The downstream problem is defined as follows:

{\vspace{\eqspace mm}\small
\begin{align}
\label{eq:downstream}
    &\min_{\Tilde{\boldsymbol{\theta}}}  \sum_{e_k \in \mathcal{E}_{test}} \mathbb{E}_{\mathbb{P}(X^{e_k}, Z^{e_k}, Y^{e_k})} \ell(f(X^{e_k}, \Tilde{\boldsymbol{\theta}}), Y^{e_k}) \\
    &\text{subject to}\quad  f(X^{e_k}, \Tilde{\boldsymbol{\theta}}) = f(T(X^{e_k}, Z^{e_k}, e_l), \Tilde{\boldsymbol{\theta}}), \nonumber \\
    & \mathbb{E}_{\mathbb{P}(X^{e_k}, Z^{e_k}), \mathbb{P}(X^{e_l}, Z^{e_l})} \big[g(X^{e_k}, Z^{e_k}) + g(X^{e_l}, Z^{e_l})\big] = 0 \nonumber
\end{align}
}
where $\mathbf{x}^{e_k}\sim\mathbb{P}(X^{e_k})$, $\mathbf{x}^{e_l}\sim\mathbb{P}(X^{e_l})$, $\mathbf{z}^{e_k}\sim\mathbb{P}(Z^{e_k})$, $\forall e_k,e_l\in\mathcal{E}_{test}$, $i\neq j$. $\boldsymbol{\theta}^{*}$ is the initialization of $\Tilde{\boldsymbol{\theta}}$.
\end{problem}

The challenges presented in \cref{prob:our-problem} arise from the need for meta-learning models. Specifically, the framework conducts meta-training across all the training domains $\mathcal{E}_{train}$, utilizing the breadth of these domains to learn a robust set of initial parameters. However, the true test of generalization and fairness occurs during the subsequent phase, where the meta parameters serve as initial parameters for downstream tasks on a limited subset of samples from the testing domains $\mathcal{E}_{test}$. This problem underscores a significant challenge: ensuring that the model not only adapts to new, unseen domains with very few examples but also maintains consistent and fair performance across the comprehensive domain set $\mathcal{E}$. The sparse availability of samples in $\mathcal{E}_{test}$ compounds this difficulty, demanding that the initial parameters derived from meta-training possess an intrinsic capability to generalize effectively and equitably, even under constrained conditions.
A key aspect of tackling this problem involves addressing how closely the data feature distributions in testing domains resemble those in the observed training domains $\mathcal{E}_{train}$. The existing methods on domain generalization \cite{zhang2022towards,robey2021model} incorporate this consideration and introduce solutions primarily focused on decomposing variations in data features across domains into distinct latent spaces. To ensure fairness, data features are categorized into sensitive and non-sensitive components. It is assumed that the dependency of sensitive features on labels might vary across domains, which may not be strictly domain-invariant or domain-specific. This nuanced understanding acknowledges that fairness levels across different domains may differ, enhancing the realism and applicability of the proposed solutions.

\begin{algorithm}[!t]
\small
    \caption{Fairness-Enhanced Meta-Learning for Domain Generalization.}
    \label{alg:our_alg}
    \begin{flushleft}
        \textbf{Require}: domain transformation model $T$. \\
        \textbf{Require}: $\{\boldsymbol{\theta}_m, \boldsymbol{\theta}_s, \boldsymbol{\theta}_c, \boldsymbol{\theta}_a, \boldsymbol{\theta}_z, \boldsymbol{\phi}_{i}, \boldsymbol{\phi}_o\}$ \\
        \textbf{Initialize}: primal and dual learning rate $\eta_p, \eta_d$, empirical constant $\gamma_1, \gamma_2$.
    \end{flushleft}
\begin{algorithmic}[1]
    \While{not done}
        \State Sample batch of tasks $\mathcal{T}=\{\mathcal{B}^{sup}, \mathcal{B}^{qry}\} \sim p(\mathcal{T})$
        \For{each $\mathcal{T}$}
            \State $\boldsymbol{\theta}^{'} = \boldsymbol{\theta}$
            \State $\mathcal{B}^{sup}_{aug} \leftarrow \{T(\mathbf{x}_i, z_i, y_i)\mid (\mathbf{x}_i, z_i, y_i)\in\mathcal{B}^{sup} \}$
            \State $\boldsymbol{\theta}^{'}\leftarrow\text{Adam}(\mathcal{L}(\boldsymbol{\theta}^{'}, \mathcal{B}^{sup}, \mathcal{B}^{sup}_{aug}),\boldsymbol{\theta}^{'}, \eta_p)$
            \State update $\lambda_1^{'}, \lambda_2^{'}$ on $\boldsymbol{\theta}', \mathcal{B}^{sup}, \mathcal{B}^{sup}_{aug}$
            \State $\mathcal{B}^{qry}_{aug} \leftarrow \{T(\mathbf{x}_j, z_j, y_j)\mid (\mathbf{x}_j, z_j, y_j)\in\mathcal{B}^{qry} \}$
            \State calculate $\mathcal{L}(\boldsymbol{\theta}^{'}, \mathcal{B}^{qry}, \mathcal{B}^{sup}_{aug})$
        \EndFor
        \State $\boldsymbol{\theta}\leftarrow \boldsymbol{\theta}-\eta_p \cdot \nabla_{\boldsymbol{\theta}} \sum_{\mathcal{T} \sim p(\mathcal{T})} \mathcal{L}(\boldsymbol{\theta}^{'}, \mathcal{B}^{qry}, \mathcal{B}^{qry}_{aug})$
        \State update $\lambda_1, \lambda_2$ on $\boldsymbol{\theta}, \mathcal{B}^{qry}, \mathcal{B}^{qry}_{aug}$
    \EndWhile
    \Procedure{$T$}{$\mathbf{x}, z, y$}
    \State $\mathbf{c}= E^{c}(E^{m}(\mathbf{x}, \boldsymbol{\theta}_{m}) ,\boldsymbol{\theta}_{c})$
    \State Sample $\mathbf{a}'\sim\mathcal{N}(0,I_a)$
    \State Sample $\mathbf{s}'\sim\mathcal{N}(0,I_s)$
    \State $\mathbf{x}'= G^o(G^{i}(\mathbf{c}, \mathbf{a}', \boldsymbol{\phi}_{i}), \mathbf{s}', \boldsymbol{\phi}_o)$
    \State $z'= h(\mathbf{a}', \boldsymbol{\theta}_z)$
    \State \textbf{return} $(\mathbf{x}', z', y)$
    \EndProcedure
\end{algorithmic}
\end{algorithm}

\begin{figure}[!t]
    \includegraphics[width=1.0\linewidth]{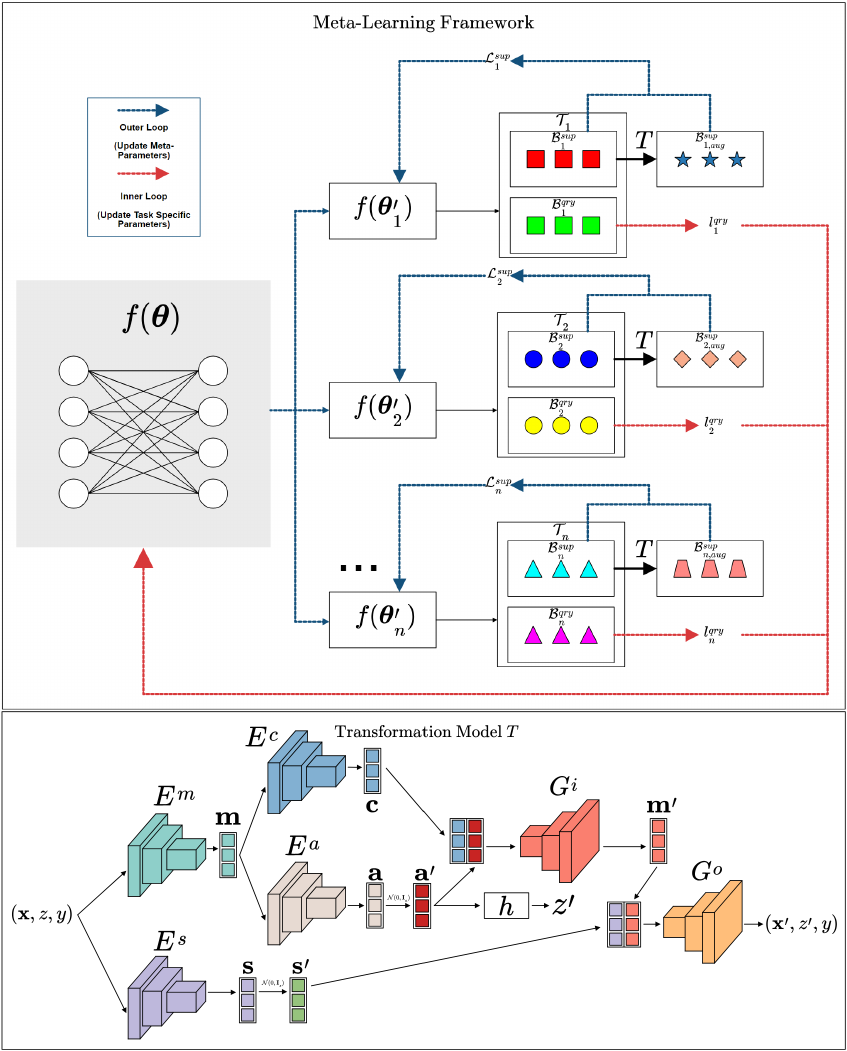}
    \caption{(Top) An overview of our framework. The red lines and the blue lines correspond to outer loop and inner loop respectively. (Bottom) The transformation model $T$. It generates an augmented example having the same content factor as the input example but has different style and sensitive factors sampled from their associated distributions that encode a new synthetic domain.}
    \label{fig:framework}
    \vspace{-4mm}
\end{figure}

\textbf{Implementation of \sysname{}.} Our proposed implementation is shown in \cref{alg:our_alg}.  In lines 14-21, we describe the $T$ procedure that takes an example $(\mathbf{x}, z, y)$ as input and returns an augmented example $(\mathbf{x}^\prime, z^\prime, y)$ from a new synthetic domain as output. The augmented example has the same content factor as the input example but has different style and sensitive factors sampled from their associated distributions that encode a new synthetic domain as shown in \cref{fig:framework}.
Line 3-10 show the inner loop updating the task specifice parameters for \sysname{}, and line 11-12 show the outer loop udating the meta parameters. In line 6 and line 8, for each example in a data batch $\mathcal{B}$, we apply the procedure $T$ to generate an augmented example from in new synthetic domain. The loss functions are defined in \cref{eq:l_cls,eq:l_inv,eq:l_fair,eq:l_total} and the \cref{eq:update_lambda1,eq:update_lambda2} show how the hyperparameters $\lambda_1, \lambda_2$ are updated.

\textbf{Classification loss} Given data batch $\mathcal{B}=\{(\mathbf{x}_i, y_i, z_i)\}_{i=1}^{|\mathcal{B}|}$ and classifier $f$ parameterized by $\boldsymbol{\theta}$, the classification loss $\mathcal{L}_{cls}(\boldsymbol{\theta}, \mathcal{B})$ is defined as:

{\vspace{\eqspace mm}\small
\begin{align}
\label{eq:l_cls}
\mathcal{L}_{cls}(\boldsymbol{\theta}, \mathcal{B}) = \frac{1}{|\mathcal{B}|}\sum_{i=1}^{|\mathcal{B}|}\ell(y_i, f(\mathbf{x}_i,\boldsymbol{\theta}))
\end{align}
}
where we use crossentropy as the distance metric for $d(\cdot)$.

\textbf{Invariance loss} With $\mathcal{B}_{aug}$ whose data points are transformed from $\mathcal{B}$ by $T$, the invariance loss $\mathcal{L}_{inv}(\boldsymbol{\theta}, \mathcal{B}, \mathcal{B}_{aug})$ is based on the difference between predictions of original and transformed data points:

{\vspace{\eqspace mm}\small
\begin{align}
\label{eq:l_inv}
    \mathcal{L}_{inv}(\boldsymbol{\theta}, \mathcal{B}, \mathcal{B}_{aug}) = \frac{1}{|\mathcal{B}|} \sum_{i=1}^{|\mathcal{B}|} d[f(\mathbf{x}_i, \boldsymbol{\theta}), f(\mathbf{x}_j, \boldsymbol{\theta})]
\end{align}
}
where $(\mathbf{x}_i, z_i, y_i)$ is a data point from the original data batch $\mathcal{B}$, and $(\mathbf{x}_j, z_j, y_j) = T(\mathbf{x}_i, z_i, y_i)$ is its transformed counterpart in $\mathcal{B}_{aug}$. We consider KL-divergence as the distance metric for $d(\cdot)$.

\textbf{Fairness loss} The fairness loss $\mathcal{L}_{fair}(\boldsymbol{\theta}, \mathcal{B}, \mathcal{B}_{aug})$ is calculated to ensure that the model’s predictions remain fairness across different sensitive attributes across source domain and synthetic domain.

{\vspace{\eqspace mm}\small
\begin{align}
\label{eq:l_fair}
    \mathcal{L}_{fair}(\boldsymbol{\theta}, \mathcal{B}, \mathcal{B}_{aug}) &= \left|\frac{1}{|\mathcal{B}|}\sum_{(\mathbf{x}_i,z_i)\in\mathcal{B}} g(f(\mathbf{x}_i,\boldsymbol{\theta}),z_i)\right| \nonumber \\
    &+ \left|\frac{1}{|\mathcal{B}_{aug}|}\sum_{(\mathbf{x}_j,z_j)\in\mathcal{B}_{aug}} g(f(\mathbf{x}_j,\boldsymbol{\theta}),z_j)\right|\\
    \textit{where} \quad g(f(\mathbf{x}_i,\boldsymbol{\theta}),z_i)&=\Big|\frac{1}{p_1(1-p_1)}\Big(\frac{z_i+1}{2}-p_1\Big)f(\mathbf{x}_i,\boldsymbol{\theta}) \Big| \nonumber
\end{align}
}
where $|\cdot|$ is the absolute function. $p_1$ is the proportion of samples in group $z=1$ and correspondingly $1-p_1$ is the proportion of samples in group $z=-1$. 

\textbf{Total loss} The overall loss function $\mathcal{L}(\boldsymbol{\theta}, \mathcal{B}, \mathcal{B}_{aug})$ is then formulated as a weighted sum of the classification, invariance, and fairness losses:

{\vspace{\eqspace mm}\small
\begin{align}
\label{eq:l_total}
\mathcal{L}(\boldsymbol{\theta}, \mathcal{B}, \mathcal{B}_{aug}) &= \mathcal{L}_{cls}(\boldsymbol{\theta}, \mathcal{B}) + \lambda_1 \cdot \mathcal{L}_{inv}(\boldsymbol{\theta}, \mathcal{B}, \mathcal{B}_{aug}) \nonumber \\
& + \lambda_2 \cdot \mathcal{L}_{fair}(\boldsymbol{\theta}, \mathcal{B}, \mathcal{B}_{aug})
\end{align}
}
where $\lambda_1$ and $\lambda_2$ are hyperparameters that balance the contributions of the invariance and fairness losses, respectively.

The task-specific model parameters are updated using the Adam optimizer, and the meta parameters are updated by using gradient descent on the sum of losses calculated on the query set of each task. Simultaneously, the dual variables $\lambda_1$ and $\lambda_2$ are adjusted to penalize any violations of the fairness and invariance constraints. The updates for these dual variables are performed as follows:

{\vspace{\eqspace mm}\small
\begin{align}
\lambda_1 &\leftarrow \text{max}\{ [\lambda_1 + \eta_d \cdot (\mathcal{L}_{inv}(\boldsymbol{\theta}, \mathcal{B}) - \gamma_1)], 0\} \label{eq:update_lambda1}\\
\lambda_2 &\leftarrow \text{max}\{ [\lambda_2 + \eta_d \cdot (\mathcal{L}_{fair}(\boldsymbol{\theta}, \mathcal{B}) - \gamma_2)], 0\} \label{eq:update_lambda2}
\end{align}
}
where $\gamma_1,\gamma_2>0$ are constants.
\begin{table*}[!t]
\tiny
    \centering
    \setlength\tabcolsep{3pt}
    \caption{Performance on the New-York-Stop-and-Frisk dataset (bold is the best; underline is the second best).}
    \vspace{-3mm}
    \label{tab:result-NY}
    \begin{tabular}{l|c|c|c|c|c|c}
        \toprule
         & \multicolumn{6}{c}{ \textbf{ Accuracy $\uparrow$ / $\Delta$DP $\downarrow$ / \textbf{$\Delta$EOPP} $\downarrow$ / \textbf{$\Delta$EO} $\downarrow$ }} \\ 
        \cmidrule(lr){2-7}
        \multirow{-2}{*}{\textbf{Methods}} & $R$ & $B$ & $M$ & $Q$ & $S$ &\textbf{Avg}\\
        \cmidrule(lr){1-1} \cmidrule(lr){2-2} \cmidrule(lr){3-3} \cmidrule(lr){4-4} \cmidrule(lr){5-5} \cmidrule(lr){6-6} \cmidrule(lr){7-7}
        ERM \cite{vapnik1999nature} & 62.97 / 0.032 / 0.019 / 0.028 & 58.83 / 0.097 / 0.072 / 0.085 & 62.37 / 0.067 / 0.048 / 0.041 & 65.01 / 0.070 / 0.077 / 0.072 & 60.78 / 0.118 / 0.079 / 0.103 & 61.99 / 0.077 / 0.059 / 0.066 \\
        IRM \cite{arjovsky2019invariant} & 58.10 / 0.018 / 0.007 / 0.015 & 56.73 / 0.081 / 0.063 / 0.071 & 60.68 / 0.038 / 0.024 / 0.030 & 62.71 / 0.014 / \underline{0.010} / 0.025 & 59.34 / 0.044 / \underline{0.007} / 0.037 & 59.51 / 0.039 / \textbf{0.022} / 0.036 \\
        GroupDRO \cite{sagawa2019distributionally} & 62.15 / 0.054 / 0.037 / 0.050 & \underline{60.08} / 0.103 / 0.080 / 0.088 & \textbf{62.87} / 0.082 / 0.067 / 0.054 & 64.55 / 0.073 / 0.079 / 0.076 & 61.64 / 0.126 / 0.088 / 0.110 & \underline{62.26} / 0.087 / 0.070 / 0.076 \\
        Mixup \cite{yan2020improve} & 63.98 / 0.030 / 0.019 / 0.027 & 56.30 / 0.073 / 0.055 / 0.065 & 60.12 / 0.054 / 0.039 / 0.032 & 65.17 / 0.033 / 0.030 / 0.037 & 62.96 / 0.101 / 0.085 / 0.086 & 61.71 / 0.058 / 0.046 / 0.050 \\
        DDG \cite{zhang2022towards} & \underline{64.78} / 0.020 / 0.010 / 0.025 & 55.84 / \underline{0.059} / 0.052 / 0.061 & 59.87 / 0.042 / 0.023 / 0.030 & 62.97 / 0.028 / 0.011 / 0.039 & 56.70 / \underline{0.039} / 0.012 / 0.037 & 60.03 / \underline{0.038} / \textbf{0.022} / 0.038 \\
        MBDG \cite{robey2021model} & 62.82 / \underline{0.003} / \textbf{0.002} / \textbf{0.002} & 57.04 / 0.076 / 0.062 / 0.068 & 61.00 / 0.046 / 0.032 / 0.023 & 63.39 / \textbf{0.001} / 0.016 / \underline{0.018} & 58.88 / 0.062 / 0.021 / 0.050 & 60.63 / \underline{0.038} / 0.026 / \underline{0.032} \\
        \cmidrule(lr){1-7}
        DDG-FC & 59.82 / 0.061 / 0.066 / 0.068 & 56.85 / \textbf{0.030} / \textbf{0.019} / \textbf{0.020} & 60.66 / \textbf{0.015} / \underline{0.016} / 0.040 & 57.55 / 0.062 / 0.054 / 0.061 & 59.95 / \textbf{0.018} / \textbf{0.002} / \textbf{0.008} & 58.97 / \textbf{0.037} / 0.031 / 0.039 \\
        MBDG-FC & 62.65 / \textbf{0.001} / 0.005 / \textbf{0.002} & 57.01 / 0.076 / 0.061 / 0.066 & 60.96 / 0.046 / 0.032 / \underline{0.022} & 63.38 / \textbf{0.001} / 0.016 / \textbf{0.017} & 58.85 / 0.063 / 0.023 / 0.050 & 60.57 / \textbf{0.037} / 0.027 / \underline{0.032} \\
        EIIL \cite{creager2021environment} & 64.36 / 0.019 / 0.021 / 0.019 & 56.10 / 0.069 / 0.038 / 0.066 & 60.46 / 0.043 / 0.019 / 0.026 & 62.82 / 0.013 / 0.019 / 0.020 & 58.08 / 0.052 / 0.021 / 0.045 & 60.36 / 0.039 / \underline{0.024} / 0.035 \\
        FarconVAE \cite{oh2022learning} & 58.70 / 0.054 / 0.032 / 0.036 & \textbf{60.80} / 0.076 / \underline{0.027} / \underline{0.039} & \underline{62.50} / 0.107 / 0.027 / 0.036 & \underline{65.30} / \underline{0.007} / 0.029 / 0.038 & 61.20 / 0.056 / 0.027 / \underline{0.031} & 61.70 / 0.060 / 0.028 / 0.036 \\
        FEDORA \cite{zhao2024algorithmic} & 63.79 / 0.036 / 0.022 / 0.037 & 59.19 / 0.132 / 0.117 / 0.113 & 61.53 / 0.139 / 0.126 / 0.110 & 62.64 / 0.019 / 0.014 / 0.023 & \underline{63.19} / 0.076 / 0.087 / 0.065 & 62.07 / 0.080 / 0.073 / 0.070 \\
        \cmidrule(lr){1-7}
        \sysname{} (Ours) & \textbf{65.81} / 0.006 / \underline{0.003} / \underline{0.006} & 57.21 / 0.076 / 0.058 / 0.066 & 60.26 / \underline{0.019} / \textbf{0.012} / \textbf{0.015} & \textbf{65.31} / 0.016 / \textbf{0.004} / \textbf{0.017} & \textbf{63.79} / 0.068 / 0.030 / 0.054 & \textbf{62.48} / \textbf{0.037} / \textbf{0.022} / \textbf{0.031} \\
        \bottomrule
    \end{tabular}
    \vspace{-3mm}
\end{table*}
\begin{table*}[!t]
\tiny
    \centering
    \setlength\tabcolsep{3pt}
    \caption{Performance on the YFCC100M-FDG dataset (bold is the best; underline is the second best).}
    \vspace{-3mm}
    \label{tab:result-YFCC100M}
    \begin{tabular}{l|c|c|c|c}
        \toprule
         & \multicolumn{4}{c}{ \textbf{ Accuracy $\uparrow$ / $\Delta$DP $\downarrow$ / \textbf{$\Delta$EOPP} $\downarrow$ / \textbf{$\Delta$EO} $\downarrow$ }} \\ 
        \cmidrule(lr){2-5}
        \multirow{-2}{*}{\textbf{Methods}} & $d_0$ & $d_1$ & $d_2$ & \textbf{Avg}\\
        \cmidrule(lr){1-1} \cmidrule(lr){2-2} \cmidrule(lr){3-3} \cmidrule(lr){4-4} \cmidrule(lr){5-5}
        ERM \cite{vapnik1999nature} & 89.69 / 0.133 / \underline{0.005} / \underline{0.007} & 86.92 / 0.049 / 0.005 / 0.017 & 87.18 / 0.050 / 0.004 / 0.007 & 87.93 / 0.077 / 0.004 / 0.011 \\
        IRM \cite{arjovsky2019invariant} & 67.05 / 0.067 / 0.015 / 0.018 & 65.80 / 0.044 / 0.009 / 0.015 & 71.01 / 0.040 / \underline{0.002} / 0.012 & 67.95 / \underline{0.050} / 0.009 / 0.015 \\
        GroupDRO \cite{sagawa2019distributionally} & 89.20 / 0.138 / \textbf{0.001} / 0.026 & 66.63 / 0.048 / 0.004 / 0.011 & 85.99 / 0.048 / 0.003 / \underline{0.002} & 80.61 / 0.078 / \underline{0.002} / 0.013 \\
        Mixup \cite{yan2020improve} & \underline{90.00} / 0.130 / \textbf{0.001} / \textbf{0.004} & 86.06 / 0.050 / 0.005 / 0.020 & 86.70 / 0.049 / \underline{0.002} / 0.007 & 87.58 / 0.076 / \underline{0.002} / 0.010 \\
        DDG \cite{zhang2022towards} & 83.74 / 0.093 / 0.032 / 0.067 & 88.26 / 0.056 / 0.016 / 0.034 & 89.95 / 0.043 / 0.004 / 0.003 & 87.32 / 0.064 / 0.018 / 0.035 \\
        MBDG \cite{robey2021model} & 85.70 / 0.136 / 0.029 / 0.024 & \underline{89.90} / 0.063 / 0.025 / 0.035 & 87.49 / \underline{0.036} / \textbf{0.001} / 0.006 & 87.70 / 0.079 / 0.019 / 0.022 \\
        \cmidrule(lr){1-5}
        DDG-FC & 86.46 / 0.108 / 0.038 / 0.046 & 89.32 / 0.067 / 0.030 / 0.038 & 88.04 / 0.058 / 0.017 / 0.012 & 87.94 / 0.077 / 0.028 / 0.032 \\
        MBDG-FC & \textbf{92.12} / \textbf{0.057} / 0.032 / 0.154 & 70.72 / 0.061 / \textbf{0.001} / \textbf{0.002} & 85.56 / 0.054 / \textbf{0.001} / 0.008 & 82.80 / 0.057 / 0.011 / 0.055 \\
        EIIL \cite{creager2021environment} & 71.56 / 0.064 / 0.040 / 0.065 & 68.96 / 0.049 / 0.009 / \underline{0.006} & 72.20 / 0.042 / \textbf{0.001} / \textbf{0.001} & 70.91 / 0.052 / 0.017 / 0.024 \\
        FarconVAE \cite{oh2022learning} & 84.80 / 0.175 / \textbf{0.001} / 0.011 & 72.60 / 0.048 / \underline{0.002} / 0.012 & 74.50 / 0.071 / 0.004 / 0.012 & 77.30 / 0.098 / \underline{0.002} / 0.012 \\
        FEDORA \cite{zhao2024algorithmic} & 87.40 / 0.139 / \textbf{0.001} / 0.010 & 89.50 / \textbf{0.020} / \underline{0.002} / 0.008 & \underline{90.00} / \textbf{0.030} / \underline{0.002} / 0.007 & \underline{88.97} / 0.063 / \textbf{0.001} / \underline{0.008} \\
        \cmidrule(lr){1-5}
        \sysname{} (Ours) & 83.96 / \underline{0.060} / \textbf{0.001} / 0.008 & \textbf{91.36} / \underline{0.033} / \textbf{0.001} / 0.009 & \textbf{92.47} / 0.038 / \textbf{0.001} / \underline{0.002} & \textbf{89.26} / \textbf{0.044} / \textbf{0.001} / \textbf{0.006} \\
        \bottomrule
    \end{tabular}
    \vspace{-3mm}
\end{table*}
\section{Experiments}
\label{sec:experiments}
We conducted a comprehensive evaluation of our proposed framework, \sysname{}, across a variety of domain generalization benchmarks that encompass both domain characteristics and sensitive attributes. In this assessment, \sysname{} was compared against 11 well-established baselines to illustrate its efficacy.
%We specifically excluded popular benchmarks such as PACS \cite{li2017deeper} and VLCS \cite{torralba2011unbiased} from our analysis because they lack sensitive attributes, which are crucial for assessing fairness in model performance.
The detailed empirical setup is outlined in \cref{sec:settings}, and the results of these experiments are discussed in \cref{sec:results}.
\subsection{Experimental Settings}
\label{sec:settings}
\textbf{Datasets.} We consider four datasets: ccMNIST, FairFace, YFCC100M-DFG, and New York Stop-and-Frisk(NYSF) to evaluate our \sysname{} against state-of-the-art baseline methods, where NYSF is a tabular dataset and the other three are image datasets. 

\textbf{(1) ccMNIST:} The ccMNIST is a domain generalization dataset derived from the MNIST dataset \cite{lecun1998gradient} by introducing color to the digits and backgrounds. This dataset features images of handwritten digits from 0 to 9, categorized into binary classes with digits 0-4 labeled as 0 and 5-9 labeled as 1, akin to the method used in ColoredMNIST \cite{arjovsky2019invariant}. The ccMNIST includes three distinct domains, each represented by a unique digit color (red, green, blue), encompassing a total of 70,000 images. Notably, each domain exhibits a varying degree of correlation between the class label and the sensitive attribute, the background color, quantified as $0.9$, $0.7$, and $0$ for the red, green, and blue domains, respectively.
\textbf{(2) FairFace:} The FairFace dataset \cite{Karkkainen_2021_WACV} comprises 108,501 images, portraying a balanced representation across seven racial groups: Black (B), East Asian (E), Indian (I), Latino (L), Middle Eastern (M), Southeast Asian (S), and White (W). For our experimental framework, each racial category is treated as a separate domain, with gender designated as the sensitive attribute and age (either $\geq$ or $<$ 50 years) as the binary class label.
\textbf{(3) YFCC100M-FDG:} This image dataset, a subset of the YFCC100M \cite{thomee2016yfcc100m}, curated by \textit{Yahoo Labs}, consists of 90,000 images selected randomly and divided into three domains based on the year of capture: prior to 1999 ($d_0$), 2000 to 2009 ($d_1$), and 2010 to 2014 ($d_2$), with each domain containing 30,000 images. The binary class label is determined by the outdoor or indoor tag of each image, while the latitude and longitude coordinates are translated into a sensitive attribute indicating whether the image was taken in North-America or outside of it.
\textbf{(4) NYSF:} The NYSF dataset \cite{Koh-icml-2021} documents police stops in New York City during 2011, focusing on whether pedestrians suspected of weapon possession were indeed carrying a weapon. The data, inherently biased against African Americans, is structured into five sub-city domains: Manhattan (M), Brooklyn (B), Queens (Q), Bronx (R), and Staten (S). Race (black or non-black) is used as the sensitive attribute in this real-world dataset.

\textbf{Baselines.}
In our evaluation, the performance of \sysname{} is benchmarked against 15 baseline methods, categorized into three distinct groups based on their primary focus and approach: \textbf{(a)} six state-of-the-art \textit{domain generalization} methods, which include ERM \cite{vapnik1999nature}, IRM \cite{arjovsky2019invariant}, GroupDRO \cite{sagawa2019distributionally}, Mixup \cite{yan2020improve}, DDG \cite{zhang2022towards}, and MBDG \cite{robey2021model}; \textbf{(b)} three advanced \textit{fairness-aware learning} methods that address variability in environments, namely EIIL \cite{creager2021environment}, FarconVAE \cite{oh2022learning}, and FEDORA \cite{zhao2024algorithmic}; and \textbf{(c)} two \textit{naive fairness-aware variants} of existing domain generalization methods, specifically DDG-FC and MBDG-FC, which are adaptations of DDG and MBDG with additional fairness constraints in \cref{eq:l_fair} integrated into their classification frameworks.
% It is important to note the setting differences in the EIIL and MBDG compared to the setting in our method. Both methods characterize domain shift by a different level of correlation between the class label and sensitive features but completely ignore the variation in data features.

\textbf{Evaluation metrics.}
We use three popular evaluation metrics to evaluate the group fairnesses of different methods:
\begin{itemize}[leftmargin=*]
    \item \textit{Difference in Demographic Parity} ($\Delta$DP) quantifies the disparity in positive prediction rates across groups:
    
    {\vspace{\eqspace mm}\small
    \begin{align*}
        \Delta \text{DP} = \left|\mathbb{P}(\hat{Y}=1|Z=-1) - \mathbb{P}(\hat{Y}=1|Z=1)\right|
    \end{align*}
    }
    A value of 0 indicates perfect fairness.
    \item \textit{Difference in Equal Opportunity} ($\Delta$EOPP) measures the difference in true positive rates between groups:
    
    {\vspace{\eqspace mm}\small
    \begin{align*}
        \Delta \text{EOPP} &= \Big|\mathbb{P}(\hat{Y}=1|Y=1, Z=-1) \\
        & - \mathbb{P}(\hat{Y}=1|Y=1, Z=1) \Big|
    \end{align*}
    }
    A value close to 0 indicates that both groups have an equal chance of receiving a positive outcome.
    \item \textit{Difference in Equalized Odds} ($\Delta$EO) captures differences across more comprehensive metrics, including both the true positive and false positive rates, ensuring no advantage is given to any group across the decision threshold:
    
    {\vspace{\eqspace mm}\small
    \begin{align*}
        \Delta \text{EO} &= \frac{1}{2} \Big( \Big|\mathbb{P}(\hat{Y}=1|Y=1, Z=-1) \\
        &- \mathbb{P}(\hat{Y}=1| Y=1,Z=1)\Big| + \Big|\mathbb{P}(\hat{Y}=1|Y=0, Z=-1) \\
        &- \mathbb{P}(\hat{Y}=1|Y=0, Z=1)\Big| \Big)
    \end{align*}
    }
    Achieving a $\Delta$EO of 0 is indicative of fair treatment across both outcomes.
\end{itemize}
\begin{table*}[!t]
\tiny
    \centering
    % \rowcolors{3}{white}{gray!15}
    \setlength\tabcolsep{8pt}
    \caption{Performance on the FairFace dataset (bold is the best; underline is the second best).}
    \vspace{-3mm}
    \label{tab:result-fairface}
    \begin{tabular}{l|c|c|c|c}
        % \hline
        \toprule
         & \multicolumn{4}{c}{ \textbf{ Accuracy $\uparrow$ / $\Delta$DP $\downarrow$ / \textbf{$\Delta$EOPP} $\downarrow$ / \textbf{$\Delta$EO} $\downarrow$ }} \\ 
        \cmidrule(lr){2-5}
        \multirow{-2}{*}{\textbf{Methods}} & $B$ & $E$ & $I$ & $L$ \\
        \cmidrule(lr){1-1} \cmidrule(lr){2-2} \cmidrule(lr){3-3} \cmidrule(lr){4-4} \cmidrule(lr){5-5}
        ERM \cite{vapnik1999nature} & \textbf{92.08} / 0.016 / 0.058 / 0.037 & 92.81 / 0.053 / 0.168 / 0.095 & 86.94 / 0.041 / 0.109 / 0.064 & \underline{91.99} / 0.087 / 0.090 / 0.070 \\
        IRM \cite{arjovsky2019invariant} & 90.78 / \textbf{0.001} / 0.022 / \textbf{0.001} & 68.41 / 0.107 / 0.104 / 0.091 & 59.31 / 0.037 / 0.074 / 0.047 & 91.20 / \textbf{0.031} / 0.113 / 0.051 \\
        GroupDRO \cite{sagawa2019distributionally} & 89.78 / 0.022 / 0.086 / 0.053 & 92.35 / 0.054 / 0.151 / 0.087 & 89.05 / 0.031 / 0.123 / 0.065 & 88.59 / 0.062 / \underline{0.060} / 0.042 \\
        Mixup \cite{yan2020improve} & 90.46 / \underline{0.003} / 0.040 / 0.021 & 92.66 / 0.047 / 0.091 / 0.055 & 89.82 / 0.021 / 0.061 / 0.031 & 89.66 / 0.054 / \textbf{0.056} / \textbf{0.039} \\
        DDG \cite{zhang2022towards} & 90.49 / 0.026 / \underline{0.009} / 0.007 & 92.55 / 0.027 / \underline{0.027} / \underline{0.016} & 89.21 / 0.051 / 0.121 / 0.067 & 89.13 / 0.047 / 0.081 / 0.047 \\
        MBDG \cite{robey2021model} & \underline{91.84} / 0.041 / 0.073 / 0.040 & 93.28 / 0.036 / 0.063 / 0.023 & 88.10 / 0.042 / 0.071 / 0.036 & 90.31 / 0.055 / 0.074 / \underline{0.041} \\
        \cmidrule(lr){1-5}
        DDG-FC  & 90.57 / 0.005 / 0.011 / 0.007 & 92.62 / \underline{0.003} / \textbf{0.013} / \textbf{0.008} & \underline{90.38} / 0.049 / 0.189 / 0.105 & 90.97 / 0.074 / 0.187 / 0.113 \\
        MBDG-FC  & 91.12 / 0.032 / 0.056 / 0.038 & 93.31 / 0.035 / 0.062 / 0.041 & 87.79 / 0.037 / 0.082 / 0.051 & 88.77 / \underline{0.032} / 0.077 / 0.049 \\
        EIIL \cite{creager2021environment} & 90.71 / 0.038 / 0.050 / 0.032 & 83.34 / 0.054 / 0.056 / 0.040 & 83.47 / \textbf{0.003} / \underline{0.045} / \textbf{0.007} & 88.33 / 0.087 / 0.141 / 0.097 \\
        FarconVAE \cite{oh2022learning} & 90.30 / 0.032 / 0.092 / 0.053 & 92.70 / 0.138 / 0.082 / 0.067 & 87.10 / 0.038 / 0.087 / 0.062 & 88.30 / 0.109 / 0.088 / 0.058 \\
        FEDORA \cite{zhao2024algorithmic} & 90.71 / \textbf{0.001} / \textbf{0.005} / \underline{0.003} & \textbf{94.72} / 0.032 / 0.156 / 0.083 & 89.35 / \underline{0.010} / \textbf{0.044} / \underline{0.023} & \textbf{92.56} / 0.035 / 0.110 / 0.059 \\
        \cmidrule(lr){1-5}
        \sysname{} (Ours) & 91.06 / 0.011 / 0.106 / 0.054 & \underline{94.07} / \textbf{0.002} / 0.079 / 0.040 & \textbf{91.81} / 0.025 / 0.062 / 0.035 & 91.82 / 0.065 / 0.066 / 0.049 \\
        \bottomrule
    \end{tabular}
    \vspace{-4mm}
\end{table*}
\begin{table*}[!t]
\tiny
    \centering
    % \rowcolors{3}{white}{gray!15}
    \setlength\tabcolsep{8pt}
    \begin{tabular}{l|c|c|c|c}
        % \hline
        \toprule
         & \multicolumn{4}{c}{ \textbf{ Accuracy $\uparrow$ / $\Delta$DP $\downarrow$ / \textbf{$\Delta$EOPP} $\downarrow$ / \textbf{$\Delta$EO} $\downarrow$ }} \\ 
        \cmidrule(lr){2-5}
        \multirow{-2}{*}{\textbf{Methods}} & $M$ & $S$ & $W$ & \textbf{Avg}\\
        \cmidrule(lr){1-1} \cmidrule(lr){2-2} \cmidrule(lr){3-3} \cmidrule(lr){4-4} \cmidrule(lr){5-5} 
        ERM \cite{vapnik1999nature} & \textbf{92.04} / 0.090 / 0.133 / 0.079 & 89.93 / 0.040 / 0.115 / 0.069 & 86.73 / 0.083 / 0.192 / 0.104 & 90.36 / 0.059 / 0.124 / 0.074 \\
        IRM \cite{arjovsky2019invariant} & 60.95 / \textbf{0.049} / \underline{0.043} / \underline{0.032} & 92.81 / \underline{0.012} / 0.050 / 0.017 & 89.25 / 0.032 / 0.128 / 0.055 & 78.96 / \underline{0.038} / \underline{0.076} / \underline{0.042} \\
        GroupDRO \cite{sagawa2019distributionally} & 89.82 / 0.078 / 0.144 / 0.076 & 90.73 / 0.031 / 0.040 / 0.030 & \underline{90.14} / 0.090 / 0.192 / 0.105 & 90.06 / 0.052 / 0.114 / 0.065 \\
        Mixup \cite{yan2020improve} & 89.13 / 0.068 / 0.072 / 0.042 & 90.19 / 0.034 / \underline{0.017} / 0.021 & 89.81 / 0.085 / 0.214 / 0.114 & 90.25 / 0.044 / 0.079 / 0.046 \\
        DDG \cite{zhang2022towards} & 86.45 / \underline{0.056} / 0.106 / 0.060 & 90.74 / 0.031 / 0.069 / 0.036 & 88.89 / 0.112 / 0.218 / 0.129 & 89.64 / 0.050 / 0.090 / 0.052 \\
        MBDG \cite{robey2021model} & 88.17 / 0.075 / 0.138 / 0.074 & 91.13 / 0.041 / 0.070 / 0.036 & 88.26 / 0.056 / 0.109 / 0.056 & 90.16 / 0.050 / 0.085 / 0.044 \\
        \cmidrule(lr){1-5}
        DDG-FC  & 89.00 / 0.107 / 0.210 / 0.130 & 90.52 / \textbf{0.007} / 0.027 / \underline{0.015} & 88.50 / 0.087 / 0.218 / 0.127 & 90.37 / 0.047 / 0.122 / 0.072 \\
        MBDG-FC  & 89.40 / 0.066 / 0.045 / 0.036 & 90.72 / 0.033 / 0.057 / 0.039 & 89.62 / 0.073 / 0.166 / 0.096 & 90.10 / 0.044 / 0.078 / 0.050 \\
        EIIL \cite{creager2021environment} & 84.77 / 0.137 / 0.122 / 0.113 & 90.19 / 0.046 / 0.064 / 0.041 & 86.46 / \underline{0.014} / \textbf{0.045} / \textbf{0.017} & 86.75 / 0.054 / 0.092 / 0.050 \\
        FarconVAE \cite{oh2022learning} & 85.30 / 0.154 / 0.092 / 0.066 & 89.50 / 0.044 / 0.087 / 0.060 & 86.80 / 0.190 / \underline{0.087} / 0.055 & 88.57 / 0.101 / 0.088 / 0.060 \\
        FEDORA \cite{zhao2024algorithmic} & 91.09 / 0.079 / 0.252 / 0.131 & \underline{93.62} / 0.020 / 0.057 / 0.032 & \textbf{92.48} / 0.097 / 0.232 / 0.127 & \textbf{92.08} / 0.039 / 0.122 / 0.065 \\
        \cmidrule(lr){1-5}
        \sysname{} (Ours) & \underline{91.47} / 0.087 / \textbf{0.032} / \textbf{0.027} & \textbf{94.10} / 0.030 / \textbf{0.004} / \textbf{0.012} & 86.89 / \textbf{0.010} / 0.105 / \underline{0.053} & \underline{91.60} / \textbf{0.033} / \textbf{0.065} / \textbf{0.039} \\
        \bottomrule
    \end{tabular}
    \vspace{-3mm}
\end{table*}
\begin{table*}[!t]
\tiny
    \centering
    % \rowcolors{3}{white}{gray!15}
    \setlength\tabcolsep{8pt}
    \caption{Performance on the ccMNIST dataset (bold is the best; underline is the second best).}
    \vspace{-3mm}
    \label{tab:result-ccMNIST}
    \begin{tabular}{l|c|c|c|c}
        % \hline
        \toprule
         & \multicolumn{4}{c}{ \textbf{ Accuracy $\uparrow$ / $\Delta$DP $\downarrow$ / \textbf{$\Delta$EOPP} $\downarrow$ / \textbf{$\Delta$EO} $\downarrow$ }} \\ 
        \cmidrule(lr){2-5}
        \multirow{-2}{*}{\textbf{Methods}} & $R$ & $G$ & $B$ & \textbf{Avg}\\
        \cmidrule(lr){1-1} \cmidrule(lr){2-2} \cmidrule(lr){3-3} \cmidrule(lr){4-4} \cmidrule(lr){5-5}
        ERM \cite{vapnik1999nature} & 98.69 / 0.793 / 0.065 / 0.046 & 97.68 / 0.393 / 0.014 / \underline{0.012} & 97.81 / 0.020 / 0.006 / \underline{0.008} & 98.06 / 0.402 / 0.028 / 0.022\\
        IRM \cite{arjovsky2019invariant} & 97.55 / 0.785 / 0.115 / 0.075 & 97.36 / 0.396 / 0.030 / 0.019 & 97.14 / 0.021 / 0.009 / 0.009 & 97.35 / 0.401 / 0.052 / 0.034\\
        GroupDRO \cite{sagawa2019distributionally} & \underline{99.03} / 0.800 / 0.085 / 0.052 & 97.97 / 0.399 / 0.023 / 0.017 & 97.63 / \underline{0.010} / 0.011 / 0.013 & 98.21 / 0.403 / 0.040 / 0.027\\
        Mixup \cite{yan2020improve} & 98.92 / 0.796 / 0.050 / 0.045 & 97.13 / 0.398 / 0.021 / 0.024 & 97.70 / 0.014 / 0.006 / \textbf{0.004} & 97.92 / 0.403 / 0.026 / 0.024\\
        DDG \cite{zhang2022towards} & 98.99 / 0.794 / 0.040 / 0.039 & 97.04 / 0.421 / 0.059 / 0.052 & 97.81 / 0.013 / 0.010 / 0.011 & 97.95 / 0.409 / 0.036 / 0.034\\
        MBDG \cite{robey2021model} & 98.87 / 0.787 / 0.036 / 0.025 & \underline{98.23} / 0.411 / 0.033 / 0.029 & \textbf{98.75} / 0.017 / 0.006 / \textbf{0.004} & \textbf{98.62} / 0.405 / 0.025 / 0.019\\
        \cmidrule(lr){1-5}
        DDG-FC & 98.40 / \underline{0.784} / 0.064 / 0.036 & \textbf{98.74} / 0.400 / 0.005 / \underline{0.012} & 97.87 / 0.023 / \underline{0.005} / 0.011 & 98.33 / 0.403 / 0.025 / 0.020\\
        MBDG-FC & 95.74 / 0.867 / 0.360 / 0.380 & 87.72 / 0.480 / 0.184 / 0.146 & 79.53 / 0.414 / 0.413 / 0.406 & 87.66 / 0.587 / 0.319 / 0.311\\
        EIIL \cite{creager2021environment} & 89.65 / 0.999 / 0.999 / 0.999 & 70.01 / 0.999 / 0.998 / 0.999 & 55.60 / 0.749 / 0.637 / 0.754 & 71.75 / 0.916 / 0.878 / 0.917 \\
        FarconVAE \cite{oh2022learning} & 94.30 / 0.797 / \textbf{0.021} / \textbf{0.011} & 86.80 / 0.405 / \underline{0.003} / 0.022 & 93.70 / 0.013 / 0.041 / 0.021 & 91.60 / 0.405 / 0.022 / 0.018 \\    
        FEDORA \cite{zhao2024algorithmic} &  96.95 / \textbf{0.736} / 0.027 / 0.021  &  98.08 / \underline{0.389} / 0.005 / \textbf{0.004}  &  96.65 / 0.013 / 0.011 / 0.021  &  97.23 / \textbf{0.379} / \underline{0.014} / \underline{0.015} \\
        \cmidrule(lr){1-5}
        \sysname{} (Ours) & \textbf{99.09} / \underline{0.784} / \underline{0.025} / \underline{0.017} & 97.81 / \textbf{0.385} / \textbf{0.001} / \textbf{0.004} & \underline{98.47} / \textbf{0.004} / \textbf{0.004} / \textbf{0.004} & \underline{98.46} / \underline{0.391} / \textbf{0.010} / \textbf{0.008}\\
        \bottomrule
    \end{tabular}
    \vspace{-3mm}
\end{table*}
\textbf{Architectures.}
In the construction of the semantic encoder $E^m$ and the content encoder $E^c$, both are designed with four strided convolutional layers, each followed by Instance Normalization \cite{he2016deep} and ReLU activation functions, as utilized in various image datasets such as ccMNIST, FairFace, and YFCC100M-FDG \cite{robey2021model, liu2017unsupervised}. The style encoder $E^s$ and the sensitive encoder $E^a$ are configured with 6 strided convolutional layers, which utilize ReLU activation, succeeded by an adaptive average pooling layer and a trio of fully connected (FC) layers. The architecture for the inner level decoder $G^i$ and the outer level decoder $G^o$ includes an upsampling layer followed by 4 convolutional layers. The sensitive classifier at the inner level incorporates an FC layer equipped with 2 neurons employing a Sigmoid activation function. The outer level discriminator $D^o$ employs a multi-scale structure as proposed by \cite{wang2018high} to ensure that $G^o$ yields realistic details and accurate global structure. In contrast, the inner level discriminator $D^i$ is composed of a straightforward FC layer with 112 neurons, activated by ReLU. The stage 2 classifier utilizes a ResNet-50 architecture \cite{he2016deep}. For the NYSF dataset, following the guidelines from \cite{oh2022learning}, all networks are exclusively formed from FC layers, including the stage 2 classifier, which comprises 4 FC layers.

\textbf{Model selection.} 
In our approach to model selection within the domain generalization framework, we adhere to the leave-one-domain-out validation criteria, a methodology supported by \cite{robey2021model} and identified as one of the three prominent methods by \cite{gulrajani2020search}. This involves evaluating \sysname{} on a training domain that is withheld during the training process and averaging the performance across the remaining $|\mathcal{E}_{train}|-1$ domains.
\subsection{Results}
\label{sec:results}
\textbf{Quantitative results.} 
For all tables in the paper, the results shown in each column represent performance on the test domain, using the rest as training domains.

Our method \sysname{} demonstrates superior performance in maintaining fairness across different datasets, significantly outperforming both traditional domain generalization methods and state-of-the-art fairness-aware approaches. For instance, in the York-Stop-and-Frisk dataset (Table~\ref{tab:result-NY}), \sysname{} achieves top fairness metrics (0\% for $\Delta$DP, 0\% for $\Delta$EOPP, and 0.1\% for $\Delta$EO) and shows a notable accuracy improvement of 0.22\% over the best baseline. This trend is consistently observed across other datasets as well.

In the YFCC100M-FDG dataset (Table~\ref{tab:result-YFCC100M}), \sysname{} not only upholds the highest fairness levels (0.6\% for $\Delta$DP, 0\% for $\Delta$EOPP, 0.2\% for $\Delta$EO) but also achieves a comparable accuracy improvement of 0.29\%. These results underline the effectiveness of \sysname{} in handling domain-specific variations while ensuring robust fairness across domains.

The datasets such as ccMNIST and NYSF further validate \sysname{}'s performance. For the FairFace dataset (Table~\ref{tab:result-fairface}), our method reports better fairness metrics (0.5\% for $\Delta$DP, 1.1\% for $\Delta$EOPP, 0.3\% for $\Delta$EO) with a slight trade-off in accuracy (0.48\% lower than the best baseline). Similarly, in the ccMNIST dataset (Table~\ref{tab:result-ccMNIST}), \sysname{} maintains competitive fairness metrics and accuracy, demonstrating its adaptability and efficiency across varying experimental settings.

Our observations indicate that \sysname{} consistently delivers strong performance on fairness metrics while maintaining competitive accuracy, affirming its potential for widespread applicability in real-world settings that demand fairness outcomes across diverse populations. This consistent performance is particularly notable in the context of challenging datasets such as York-Stop-and-Frisk and YFCC100M-FDG, where \sysname{} excels in achieving top-tier results in fairness, a critical quality for models deployed in sensitive applications.

The analysis extends to datasets like ccMNIST and NYSF, \sysname{} shows only marginal discrepancies in accuracy, yet continues to uphold superior fairness metrics. This ability to balance fairness with accuracy underpins the versatility of \sysname{}, making it a robust solution for scenarios that extend beyond traditional domain applications. Moreover, the integration of \sysname{} with domain-specific requirements showcases its adaptability and readiness to tackle the intrinsic variability and unpredictability of real-world data.
% Unlike traditional domain generalization approaches that may falter under the complex dynamics of real-world applications, \sysname{} is designed to thrive amidst these challenges, leveraging its foundational meta-learning strategy to enhance both generalization and fairness across domains.

In conclusion, \sysname{} stands out as a formidable framework in the landscape of domain generalization and fairness-aware meta-learning, offering significant improvements over both conventional and state-of-the-art methods. Its dual strengths in maintaining high classification accuracy while excelling in fairness across varied domains position \sysname{} as a transformative tool for deploying robust and fairness model in diverse real-world settings. This generalizability, coupled with the method's inherent flexibility to adapt to various data characteristics and domain shifts.
\begin{figure*}[!t]
    \centering
    \includegraphics[width=0.9\linewidth]{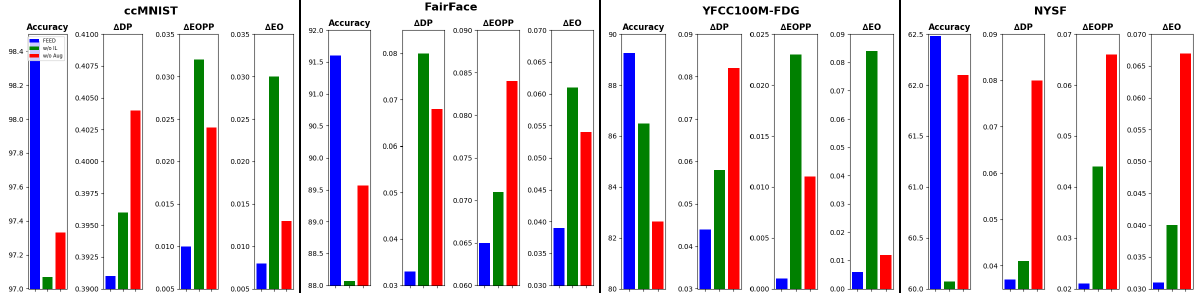}
    \vspace{-3mm}
    \caption{Ablation study on four datasets. Results are plotted as averages across all domains.}
    \label{fig:abs}
    \vspace{-5mm}
\end{figure*}
\textbf{Ablation studies.} We conducted two ablation studies.
(1) The difference between the \sysname{} and the first ablation study (\sysnameabs) is that the latter does not update the task-specific parameters based on the support set for the inner loop. In other words, the meta-parameters are directly updated based on the query loss which is calculated based on the meta-parameters. Without updating the task-specific parameters, it makes the ablation study hard to train good initial parameters, leading to poor generalization performance. Experimental results show that the first ablation study performs worse than \sysname{} on all four datasets on both accuracy and fairness metrics.
(2) The second study (\sysnameabss{}) does not use the transformation model $T$ to generate augmented support set and augmented query set. The parameters are updated only based on the support set and the query set. Similar to Abs1, without generating the augmented support set and the augmented query set in synthetic domains, it is much harder to learn good initial parameters. Our results demonstrate that Abs2 performs worse on all the datasets. We include the performance of such ablation studies in \cref{fig:abs}.

\section{Conclusion}
In this paper, we have introduced a novel framework for fairness-aware meta-learning aimed at enhancing domain generalization across diverse environments. By disentangling latent factors into content, style, and sensitive vectors, our approach ensures that the fairness, even in the face of domain shifts. The proposed fairness-aware invariance criterion plays a crucial role in maintaining fairness across different domains.

Our extensive experimental evaluation demonstrates that the proposed method not only achieves superior accuracy but also significantly improves fairness compared to existing state-of-the-art approaches. These results underscore the importance of incorporating fairness considerations into domain generalization frameworks.

Future work will explore the extension of our framework to handle multiple sensitive attributes and its application to more complex, real-world datasets. We aim to investigate the integration of our method with other fairness-aware learning paradigms to further enhance its fairness and generalizability.

\end{document}